\documentclass[twoside]{article}
\usepackage{amsmath}
\usepackage{mathrsfs}
\usepackage{bm}
\usepackage{graphicx}
\usepackage{amsfonts} 
\usepackage{wrapfig}
\usepackage{hyperref}

\DeclareMathOperator*{\argmin}{argmin}

\usepackage[preprint]{aistats2026}
\usepackage[round]{natbib}

\begin{document}

%

%

\twocolumn[

\aistatstitle{Scientific Data Compression and Super-Resolution Sampling}

\aistatsauthor{ Minh Vu \And Andrey Lokhov }

\aistatsaddress{ Theoretical Division \\
  Los Alamos National Laboratory \\
  Los Alamos, NM 87545 
  \And  
  Theoretical Division \\
  Los Alamos National Laboratory \\
  Los Alamos, NM 87545 } ]

\begin{abstract}
   Modern scientific simulations, observations, and large-scale experiments generate data at volumes that often exceed the limits of storage, processing, and analysis. This challenge drives the development of data reduction methods that efficiently manage massive datasets while preserving essential physical features and quantities of interest. In many scientific workflows, it is also crucial to enable data recovery from compressed representations—a task known as super-resolution—with guarantees on the preservation of key physical characteristics. A notable example is checkpointing and restarting, which is essential for long-running simulations to recover from failures, resume after interruptions, or examine intermediate results. In this work, we introduce a novel framework for scientific data compression and super-resolution, grounded in recent advances in learning exponential families. Our method preserves and quantifies uncertainty in physical quantities of interest and supports flexible trade-offs between compression ratio and reconstruction fidelity.
\end{abstract}

\section{Introduction}
The accelerating pace of scientific discovery has led to a deluge of data from simulations, experiments, and observational platforms. In domains such as climate modeling, high-energy and nuclear physics, astrophysics, fluid dynamics, and materials science, modern workflows routinely produce petabytes of data \citep{peterka2019ascr}. However, these datasets often far exceed available resources for storage, transmission, and downstream analysis. One of the notable differences between scientific and conventional image data lies in the known physical context: scientific images must preserve physically meaningful features (such as correlations or energy), while the quality of conventional images is prioritized through visual quality and human perception. The challenge of data reduction—efficiently compressing scientific data while preserving essential physical features—has emerged as a central problem across the computational science community. Development of rigorous data-reduction techniques that learn key physical characteristics and possess data-reconstruction ability with quantified uncertainties was highlighted as one of the grand research priorities by the U.S. Department of Energy's Office of Science workshops \citep{klasky2021data}.

This ability of recovering high-quality data data from compressed representations while is critical for solving down-stream inference and computational tasks that require processing of full data samples. One canonical use case of relevance to many scientific fields such as radiation transport, climate modeling, and plasma physics  \citep{bowers2004maximum,hanssen2001radar,chen2021unsupervised,matsekh2020learning} is checkpointing and restarting in high-performance computing simulations. This mechanism enables recovery from failures by periodically saving a comprehensive snapshot of the simulation states \citep{plank1994libckpt}, but storage of massive datasets produced at extreme scales presents significant input-output challenges \citep{sancho2004feasibility,ferreira2014accelerating}. Another example is Bayesian inference in inverse problems or denoising tasks \citep{gondara2016medical,durieux2020budd,hulbert2019similarity}, where full-scale samples recovered from reduced data are needed to support statistically valid inference. In both cases, recovering data that preserves domain-relevant physical quantities of interest (QoIs) is crucial. Lossless compression techniques naturally solve the problem by preserving full information, but typically yield modest compression ratios that are insufficient for cutting-edge applications \citep{son2014data}. In contrast, lossy compression methods \citep{ferreira2014accelerating,tiwari2014lazy,garg2018shiraz,calhoun2019exploring} have shown promise in achieving orders of magnitude higher compression ratios for a number of scientific applications (including climate, chemistry, combustion) \citep{baker2016evaluating,roe2022quantifying,ding2025reduced}, but they typically offer no guarantees that QoIs—such as conserved quantities or physical constraints—are preserved in reconstruction. In the field of computer vision, the concept of reconstruction of high-resolution images from compressed low-resolution ones is known as \emph{super-resolution} \citep{freeman2002example}. However, optimal algorithms for compression and recovery of scientific data obeying physical constraints have received far less attention and are currently lacking. While in the field of computer vision various perceptual quality metrics exist in order to evaluate the effectiveness of image super-resolution, in scientific applications we are specifically concerned with physical laws and constraints which our samples obey and our models must learn \citep{karniadakis2021physics}.

This challenge motivates a key question: How can we perform data reduction in a way that preserves the ability to reconstruct data from compressed representations while statistically conserving physical quantities of interest? In order to answer this question, we put forward the concept of a \emph{physics-informed super-resolution sampling}. Our approach build on recent advances in learning of exponential families, where the sufficient statistics are chosen to align with the QoIs that we wish to preserve. Recent rigorous analysis of algorithms such as Interaction Screening for discrete data and Score Matching for continuous data show that these models can be efficiently learned from scientific data with quantified uncertainty. Conceptually, this approach offers an optimal compression scheme, where data is fully reduced to learned parameters of maximum-entropy distributions. Samples from this model are guaranteed to preserve chosen physical properties encoded as sufficient-statistics of the model. The down-side of this representation is that it is decoupled from a sampling procedure: non-trivial energy-based models in high dimensions are difficult to sample from, especially using Markov Chain Monte Carlo (MCMC) methods that may exhibit exponentially large mixing times.

\begin{figure}[!htb]
    \centering
    \includegraphics[width=0.93\linewidth]{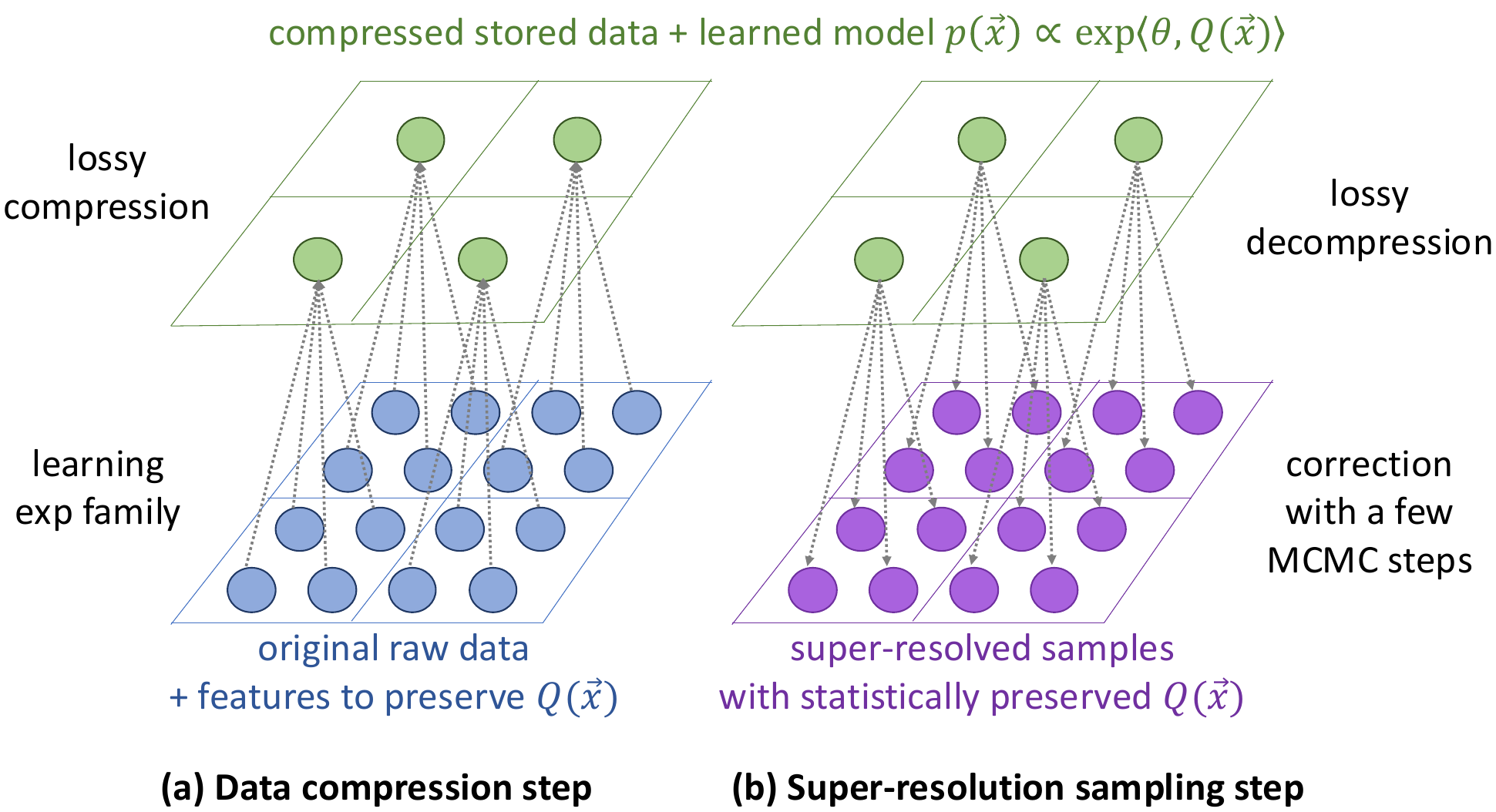}
    \caption{\small A schematic representation of our proposed approach. (a) The compression step involves learning a compact representation of the data distribution by learning a model in the exponential family with conserved quantities of interest $Q(\vec{x})$, where the desired QoIs are used as sufficient statistics. We separately store a compressed version of the original data using lossy compression. (b) When needed, this compressed data is decoded to initialize sampling procedures from the learned model, enabling efficient local correction of the data distribution and leading to a correct statistics of QoIs $Q(\vec{x})$.}\vspace{-0.25cm}
    \label{fig:enter-label}
\end{figure}

To facilitate efficient reconstruction, instead of discarding data samples after the learning procedure, we separately store a compressed version of the original data using a lossy compression method. When needed, this compressed data is decoded to initialize sampling procedures by warm-starting MCMC chains based on learned model, enabling correction of the data distribution and conservation of QoIs. This workflow greatly reduces the computational overhead while maintaining the statistical fidelity of the reconstructed data. This simple yet effective strategy leverages the fact that training samples already lie near the learned distribution and can dramatically reduce sampling costs, while maintaining statistical fidelity of the recovered data. Our approach is summarized in Figure~\ref{fig:enter-label}. Below, we study this framework for both discrete and continuous data in controlled settings, and apply it to representative scientific machine learning use cases: samples from an analog quantum computer and aluminum stress-test data from high-fidelity molecular dynamics codes. We empirically demonstrate that only a few correction steps with MCMC are sufficient to recover statistically accurate reconstructions that preserve physical QoIs. This approach bridges the gap between principled statistical modeling and practical requirements in scientific computing, offering a scalable, QoI-aware method for data reduction and reconstruction in data-intensive science.


\section{Methods}

In this section, we describe all the elements of our approach in detail: concept of exponential families and sufficient statistics; learning methods; methods for compression and decompression; and, finally, our rationale for the proposed super-resolution sampling.

\subsection{Exponential Families and Sufficient Statistics}

With the goal of learning data distributions enabling subsequent super-resolution preserving physical constraints $Q(\vec{x})$, we focus on the \textit{exponential family} distributions \citep{wainwright2008graphical}, which can represent any positive probability measures:
\begin{equation}
    P(\vec{x}) = \frac{1}{Z} \exp(E(\vec{x})),
    \label{eq:exp_family}
\end{equation}
where $\vec{x}$ is a collection of random variables, $E(\vec{x})$ is usually known as the \textit{energy function} of the model, and $Z$ is the normalization factor called the \textit{partition function}. Given a \textit{sufficient statistic} of interest $Q(\vec{x})$, the most general exponential family distribution informative of $Q(\vec{x})$ is given by the energy function parametrized by a vector of parameters $\vec{\theta}$:
\begin{equation}
    E(\vec{x}) = \vec{\theta} \cdot Q(\vec{x}).
    \label{eq:energy_function}
\end{equation}
For example, if the sufficient statistic of interest $Q(\vec{x})$ represents the first and second moments of continuous-valued data (single and two-point correlation functions $\langle x_i \rangle$ and $\langle x_i x_j \rangle$), then the exponential family is given by a multivariate Gaussian distribution, where $\vec{\theta} = (\vec{\mu}, \Sigma)$ is composed of the means $\vec{\mu}$ and the covariance $\Sigma$.
For the binary data $\sigma_i \in \{0,1\}$, the maximum entropy general model of the first and the second moment of the data (magnetizations $\langle \sigma_i \rangle$ and pair-correlation functions $\langle \sigma_i \sigma_j \rangle$) is given by the celebrated Ising model of statistical physics, with the energy function $E_{\text{Ising}}(\vec{\sigma}) = \sum_{i} h_i \sigma_i + \sum_{i < j} J_{ij} \sigma_i \sigma_j$, where the exponential family's natural parameters $\vec{\theta} = (\vec{h}, \vec{J})$ are given by the local fields and the matrix of pairwise couplings. To maintain consistency, we will use $x$ for continuous variables and $\sigma$ for discrete variables throughout the paper.

The multivariate probability distributions defined by the sufficient statistic $Q(\vec{x})$ are often referred to as Markov random fields \citep{moussouris1974gibbs}, Gibbs distributions \citep{mezard2009information}, or undirected \textit{graphical models}, where the graph structure encodes the structure of conditional dependencies of $E(\vec{x})$ \citep{wainwright2008graphical}. The task of reconstructing the parameters $\vec{\theta}$ of the probability distribution from data samples $\{\vec{x}^{(k)}\}_{k=1,\ldots,M}$ is commonly referred to as \textit{learning} of graphical models. This type of reconstruction of the parameters $\vec{\theta}$ from data presents an advantage compared to the black-box models, because it gives an explicit form of the data distribution $P(\vec{x})$ that can be used in subsequent inference tasks. Moreover, the learned vector of parameters $\vec{\theta}$ provides a compact representation of the data: for $N$-dimensional data $\vec{x} \in \mathbb{R}^{N}$, if we are only interested in moments of order at most two, such as the correlation function $x_i x_j$, only $O(N^2)$ parameters are required for a full characterization of the distribution. The down-side of this representation is that it is decoupled from a sampling procedure: non-trivial functions $E(\vec{x})$ are known to be difficult to sample.

In this paper, we propose to learn energy functions containing the QoIs as sufficient statistics, which will result in a maximum entropy distribution which conserve of these quantities of interest. In what follows, we discuss our approach to learning exponential family distributions from data, and our solution for circumventing the sampling issue.

\subsection{Learning of Exponential Families}

Reconstruction of parameters $\vec{\theta}$ in the energy function \eqref{eq:energy_function} is not a trivial task in both discrete and continuous settings. For instance, popular methods such as maximum likelihood are intractable due to the difficulty of evaluation of the partition function $Z$ which in general has exponential complexity in the dimension of the random variables \citep{cooper1990computational}. Provably polynomial-time algorithms for consistent exponential family recovery have only been established recently, see e.g. \citep{bresler2015efficiently,vuffray2016interaction,vuffray2020efficient,koehler2022statistical,pabbaraju2023provable}.

\paragraph{Discrete graphical models.} In discrete variable settings, we will use the Generalized Regularized Interaction Screening Estimator (GRISE) introduced in \citep{vuffray2020efficient} which is applicable to learning of all graphical models with finite-alphabet variables and multi-body interactions, and nearly matching optimal sample-complexity in cases where information-theoretic bounds are known. In this paper, for simplicity, we will focus on a special subclass of binary variable with pairwise interactions, known as Ising models \citep{vuffray2016interaction}. In its simplest form, GRISE for Ising models is based on minimizing a convex local loss function -- the Interaction Screening Objective (ISO): $(\hat{J}_i,\hat{h}_i) = \argmin_{(J_i,h_i)} S_i(J_i,h_i)$, where
\begin{align}
    S_i = \frac{1}{M}\sum_{m=1}^M \exp \Big( -\sum_{j\neq i} J_{ij}\sigma_i^{(m)} \sigma_j^{(m)} - h_i \sigma_i^{(m)} \Big).
\end{align}
Symmetrized estimated couplings that are sufficiently small to zero can be thresholded, thus recovering the graph structure in the cases where the graph is sparse.

\paragraph{Continuous exponential families.} In the case of Gibbs distributions with continuous variables, we use Score Matching \citep{hyvarinen2005estimation} as the algorithm of choice. The idea behind Score Matching is to bypass the intractability of the normalization constant $Z_{\theta}$ by exploiting the gradient of the distribution with respect to data $\vec{x}$. The gradient of the log-likelihood is called the score function, $ \nabla \log p_{\theta}(x) =: \psi_{\theta}(x)$. The parameters $\vec{\theta}$ of the exponential family distributions can be estimated by defining the following score matching objective: Given a data distribution $p_d(x)$ and an approximating distribution $p_{\theta}(x)$ with parameters, the score matching objective is defined as
\begin{align}
    J(\theta) = \frac{1}{2} \int p_d(x) \| \psi_{\theta}(x) - \psi_{d}(x) \|^2dx.
\end{align}
The consistency properties of Score Matching have been rigorously established in recent works \citep{koehler2022statistical, pabbaraju2023provable} for exponential families with polynomial energy functions.

\subsection{Sampling from the learned Graphical Model}

Once the energy function of the model is learned, in principle, one can use Markov Chain Monte Carlo (MCMC) methods to produce independent samples from the resulting density.

\paragraph{Glauber dynamics for discrete Gibbs distributions.} Glauber dynamics \citep{glauber1963time} is a well-known method for sampling from the Ising model and closely related to the Gibbs sampling procedure. The process begins with an arbitrary initial configuration $\bm{\sigma}^{t=0} \in \{-1, +1\}^p$, which is often chosen at random. At each time step, a single node $\sigma_i$ is selected uniformly at random for updating according to the conditional probability of the variable. For example, for Ising model with zero local fields, the update takes on value $+1$ with probability
\begin{align}
    \mathbb{P} (\sigma_i^{t+1} = +1 | \sigma_{j\neq i}^{t}) = \frac{\exp \big( 2\sum_{j\neq i} J_{ij}\sigma_j^t \big)}{1 + \exp \big(  2\sum_{j\neq i} J_{ij}\sigma_j^t \big)}
\end{align}
and is $-1$ otherwise. Notably, each spin update depends only on neighboring spins. It can be shown that the Gibbs distribution \eqref{eq:exp_family} is stationary with respect to the Glauber dynamics. If the dynamics quickly approaches equilibrium -- that is, when the mixing time of the model is small -- they can be used to simulate i.i.d. samples from \eqref{eq:exp_family}. While some Markov chains rapidly mix to their equilibrum distributions, it is well-known that in many other cases, the existence of ``bottlenecks'' between transition states may lead to slow mixing, oftentimes exponential in the dimension.

\paragraph{Langevin dynamics for continuous exponential family models.} In the case of continuous models, one can use Langevin dynamics for producing samples from the learned data distribution $p(x)$. In the discretized time version, the Markov chain is initialized from an arbitrary point, and then is iterated as follows: 
\begin{align}
    x_{t+1} = x_t + \epsilon \nabla_x \log p(x) + \sqrt{2\epsilon}z_t,
\end{align}
where $x_t$ are the samples drawn from the procedure at each iteration, $t = 0,1,...,T$, $\nabla_x \log p(x)$ is the gradient of the log density which gives us a step in the direction of the gradient, $z_t \sim \mathcal{N}(0,I)$ denotes the injected noise, and $\epsilon$ is a scaling factor that lets us control the magnitude of the step in the gradient direction. Similarly to Glauber dynamics for discrete models, Langevin dynamics is guaranteed to mix to the equilibrium distribution \eqref{eq:exp_family}, albeit at an exponential time for many families of models with non-trivial energy functions \citep{wainwright2008graphical}.

\subsubsection{Super-Resolution Sampling with Decompressed Data Initialization}

Usually learning and sampling problems are considered in separation. In typical learning settings, data is often discarded after the model is learned (see, e.g., score-based generative model \citep{song2019generative,song2020score}), whereas the resulting model can have a slow mixing time. We argue that while the Markov chains may suffer from convergence issues when initialized from a random initial condition, the situation is drastically different if we are able to initialize several MCMC dynamics starting with independent samples representative of the distribution $E(\vec{x})$.
\begin{figure}[h]
   \vspace{-0.3cm}
  \begin{center}
    \includegraphics[width=0.45\textwidth]{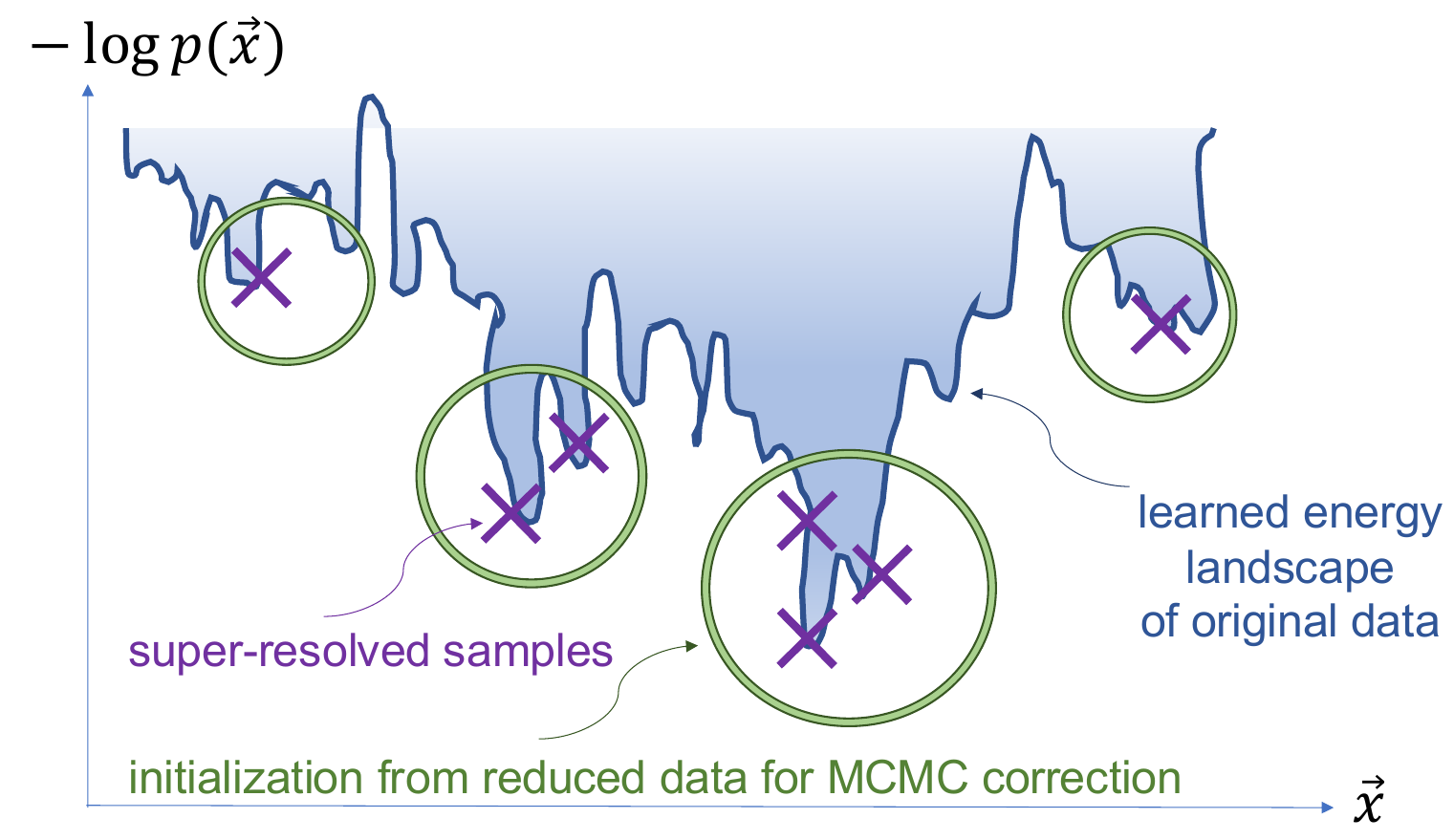}\vspace{-0.2cm}
  \end{center}
  \caption{\small Intuition behind our super-resolution sampling proposal: a stored reduced sample enables initialization of a MCMC method in the vicinity of the original sample, which then samples in a local part of the phase space based on the learned energy function $E(\vec{x})$. In practical situations, this approach may not suffer from the slow mixing of MCMC sampling starting from random initial conditions, and highlight the value of the stored reduced data.}\vspace{-0.1cm}
  \label{fig:Intuition}
\end{figure}
The idea of data-based initialization has been used in the empirical machine learning literature for a long time as a means of overcoming expensive training, for example as a part of the mechanics of ``contrastive divergence'' training for energy-based methods and other approximations to Maximum Likelihood Estimation \citep{hinton2002training,xie2016theory,nijkamp2019learning,nijkamp2020anatomy}. In theoretical settings, data-based initialization was recently shown to be effective in improving sampling of a large class of models. A direct analysis along these lines was done in \citep{koehler2023sampling} in the case of a mixture of strongly log-concave distributions supported on multiple well-separated clusters. The recent work \citep{koehler2024efficiently} provided a more general analysis for exponential families, showing provable mixing benefits of initialization with original samples.

In our setting of data compression and recovery, the original data samples are not available. However, drawing inspiration from the data-based initializations, our proposal consists in finding maximum data compression rate under which decompressed samples are still sufficiently localized in different regions of the space of configurations, and can be useful to warm-start Markov chains. The rationale for this principle is illustrated in Fig.~\ref{fig:Intuition}. Repeating this procedure using several stored reduced noise samples thus enables a physics-informed super-resolution, i.e., sampling from $P(\vec{x})$ which automatically enforces the quantities of interest $Q(\vec{x})$. An exact lossy compression and recovery scheme used is not important, but we discuss a popular choice that we adopt in this work next.

\subsection{Lossy Data Compression and Recovery}

Given a set of independent original samples, we consider the lossy compression scheme based on discrete cosine transform (DCT) \citep{ahmed2006discrete}, which is behind one of the most popular digital data compression and processing algorithms \citep{rao2014discrete} (such as JPEG and HEIF). DCT converts each sample $x_n$ from the spatial domain into the frequency domain, represented by coefficients of basis functions 
\begin{align}
    X_k = \sum_{n=0}^{N-1} x_n \cos\Big[\frac{\pi}{N}(n+\frac{1}{2})k\Big], \,\,\, k=0,...,N-1.
\end{align}
The inverse of DCT, which is simply DCT multiplied by $2/N$, converts the data back to spatial domain, which we use to for lossy recovery of samples and for subsequent initialization of Markov chains in the super-resolution sampling scheme. Multidimensional variants of the various DCT types follow straightforwardly from the one-dimensional definitions: they are simply a separable product of DCTs along each dimension. 
The inverse of a multi-dimensional DCT is again a separable product of the inverses of the corresponding one-dimensional DCTs. The premise of DCT method is that in many cases, one can reconstruct the data very accurately from only a few coefficients. The DCT coefficients, which contain up to the desired threshold of energy in the image, \vspace{-0.1cm}
\begin{equation}
\begin{aligned}
&\max_{J} && J \vspace{-0.2cm} \\
&\,\,\text{s.t.} && \Big(\sum_{k=1}^{J} X_k^2\Big)\big/\|X\|^2 \le E_{\text{presv}},\quad E_{\text{presv}}\in[0,1]
\end{aligned}
\end{equation}
are stored as the compressed data, and the remaining coefficients are disregarded. By adjusting the $E_{\text{presv}}$, we can adjust the level of compression, defined as $C = 1-E_{\text{presv}}$ . The higher $E_{\text{presv}}$ results in lower level of compression $C$, and as a result less computational effort needed to correct the distribution, leading to fast reconstruction of high quality data.
A direct application of the above equations would require $\mathcal{O}(N^2)$ operations; however, DCT can also be computed using Fast Fourier Transforms (FFT) with $\mathcal{O}(N\log N)$ operations. In our numerical experiments below, we use the DCT implementation through FFTW.jl package in Julia, connected to the FFTW library \citep{FFTW05} (a C subroutine library for computing the Discrete Fourier Transform in high dimensions).

\begin{figure*}
    \centering
    \includegraphics[width=0.99\linewidth]{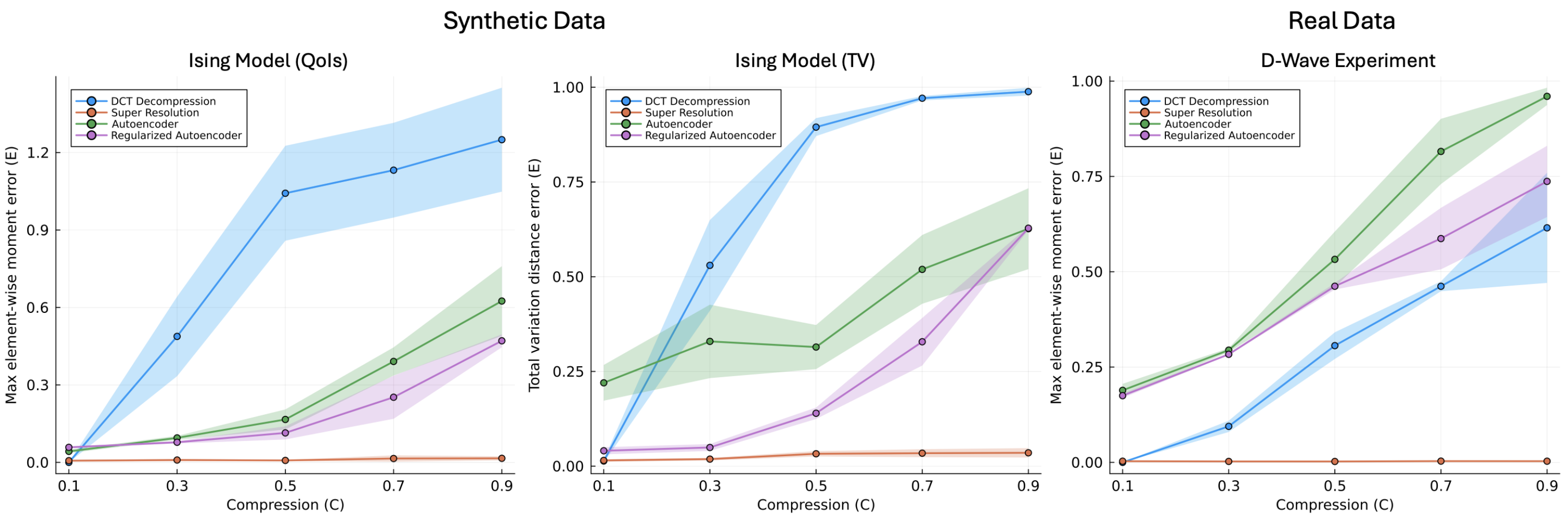}
    \caption{\small Maximum element-wise errors of the first and second moments of reconstructed samples computed across different compression levels in synthetic (left) and real (right) datasets with \emph{discrete data}. Results are averaged over 5 randomly generated models and experiments. The maximum element-wise error means and standard deviations are shown for 4 scenarios: (i) vanilla DCT decompressed samples, (ii) after application of our super-resolution correction (with 10, 12, and 10 MCMC steps for Ising-QoIs, Ising-TV, and D-Wave, respectively), (iii) reconstruction from standard autoencoders, and (iv) reconstruction from regularized autoencoders.}\vspace{-0.25cm}
    \label{fig:Ising_Dwave}
\end{figure*}
\section{Results}
In this section, we present a series of numerical experiments to evaluate our proposed approach. We test on synthetic Ising models (small enough for exact sampling), real data from an analog quantum computer, multivariate normal distributions, synthetic data from lattice field theory (generated via Langevin dynamics), and scientific data from high-fidelity materials simulations. All experiments were run on an M2 MacBook Pro (32 GB RAM) with a one-hour time limit per run. 
Our experiments follows the same setups: Given a dataset, we (i) learn an exponential-family model (with GRISE or score matching) consistent with the recorded QoIs; (ii) compress samples via DCT with varying compression levels $C\in\{0.1,0.3,\ldots,0.9\}$; and (iii) decompress the samples and apply MCMC correction (using Glauber or Langevin dynamics guided by the learned model) until the reconstructed QoIs match the stored QoIs within a desired tolerance. Each experiment is repeated for 5 times with randomly generated models and datasets, and the averaged results are reported. As a benchmark, we include comparisons with (a) a standard autoencoder trained to minimize reconstruction error and (b) a regularized autoencoder that adds a QoI penalty to the loss. More detailed information of our implementation, codes, and data are provided in the Supplementary Materials.



\subsection{Experimental Results on Discrete Data}
\textbf{Ising model:} 
We first test our workflow on a discrete Ising model defined on a two-dimensional lattice, with $p=16$ nodes, randomly chosen coupling and zero local fields. We generate $M=10^6$ i.i.d samples from the model by computing exact probabilities of all $2^{16}$ possible configurations. The goal is to compress this dataset while preserving key statistical QoIs—in this case selected to be the first and second moments of the discrete variables, i.e., $m_1(\bm{\sigma})= \frac{1}{M}\sum_{m=1}^M\bm{\sigma}^{(m)}$ and $m_2(\bm{\sigma}) = \frac{1}{M-1}\sum_{m=1}^M (\bm{\sigma}^{(m)} - m_1(\bm{\sigma})) (\bm{\sigma}^{(m)} - m_1(\bm{\sigma}))^\top$. The error tolerance of reconstruction is set to 0.05, measured as the maximum element-wise error in the first and second moments $e_1 = \|m_1(\bm{\sigma}) - m_1(\bm{\tilde{\sigma}})\|_{\infty}$ and $e_2 = \|m_2(\bm{\sigma}) - m_2(\bm{\tilde{\sigma}})\|_{\max}$, where $\bm{\tilde{\sigma}}$ are the reconstructed samples. Figure \ref{fig:Ising_Dwave} (left) illustrates the QoI errors before and after correction, showing that across all tested compression levels, applying 10 correction steps substantially reduces the errors—underscoring the effectiveness of our hybrid approach. Our analysis of the number of correction steps required to reach the prescribed error threshold indicates that more aggressive compression generally demands additional correction steps; however, the overall impact of compression on correction performance is minimal. Even under high compression, 10 correction steps are typically sufficient to recover the QoIs within the specified tolerance. For comparison, Markov chains with random initialization require roughly 250 steps to produce samples matching the target QoIs within the same tolerance. Additionally, Figure \ref{fig:Ising_Dwave} (left) shows that both standard and regularized autoencoders exhibit rapidly increasing errors as compression intensifies, while our proposed super-resolution method consistently maintains near-zero error across all compression levels. In addition to QoIs errors, we assess our framework using Total Variation (TV) distance—a more comprehensive measure directly comparing probability distributions. Figure~\ref{fig:Ising_Dwave} (middle) shows the TV distance between reconstructed data before and after correction and the ground-truth distribution, for varying levels of compression. In this example, 12 correction steps are  sufficient to reduce TV distance below 0.05 across all compression levels, while randomly initialized Markov chains require over 250 steps to achieve a comparable level of accuracy. The results demonstrate that our approach generalizes well to more stringent distributional metrics beyond QoIs errors. In what follows, we focus on preservation of QoIs as an easily computable success metric.

\begin{figure*}
    \centering
    \includegraphics[width=0.99\linewidth]{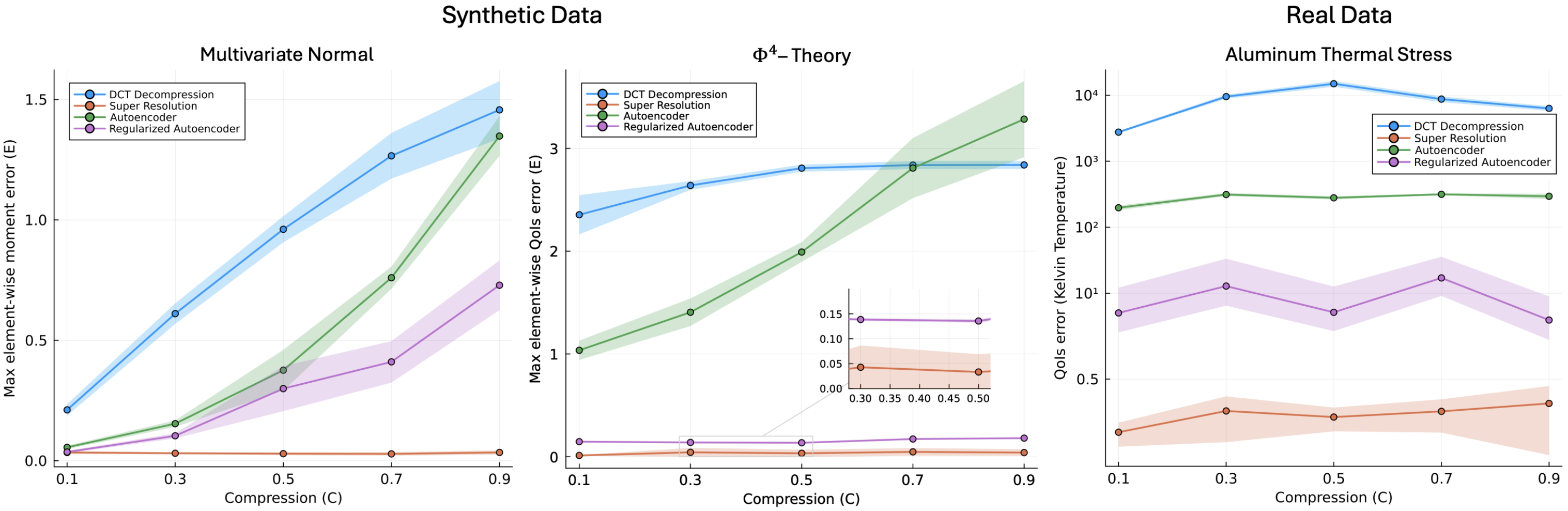}
    \caption{\small Maximum element-wise QoIs errors of reconstructed samples computed across different compression levels in synthetic (left) and real (right) datasets with \emph{continuous data}. Results are averaged over 5 randomly generated models and experiments. We use the same method as in Figure~\ref{fig:Ising_Dwave}. In our super-resolution correction part, we use 7, 20, and 50 correction steps for the Multivariate Normal, $\Phi^4$-theory, and Aluminum design experiments, respectively.}\vspace{-0.2cm}
    \label{fig:Multinormal_Phi4}
\end{figure*}

\textbf{D-Wave analog quantum computer:}
Building on the previous controlled settings, we now evaluate our framework on real data from a D-Wave 2000Q quantum annealer, an analog quantum computer that performs quantum annealing by evolving an initial state toward a ground state of an encoded Ising Hamiltonian \citep{bunyk2014architectural}. Due to thermal noise, temperature rescaling, and intrinsic device biases, the effective model from which the samples are produced is significantly distorted and does not exactly correspond to the input Ising model \citep{vuffray2022programmable, tuysuz2025learning}. Thus, the actual couplings and fields on the chip remain unknown. Using $4\times 10^6$ samples from 16 qubits arranged in the Chimera graph topology, we test our approach in a setup analogous to the synthetic Ising case, compressing the data while seeking to preserve first- and second-order correlations as QoIs. Unlike the synthetic scenario, these QoIs no longer fully characterize the underlying distribution. Figure \ref{fig:Ising_Dwave} (right) presents the QoI errors from decompressed samples and after correction, again showing significantly reduction in the QoI error, across multiple compression levels, when only 10 correction steps is applied. Remarkably, the QoIs are conserved to a great accuracy, despite the true model containing higher-order moments as sufficient statistics.

\subsection{Experiments with Continuous Data} 
\textbf{Multivariate normal distribution:}
To test our approach in the continuous setting, we use a 16-dimensional Normal distribution ${\mathcal {N}}({\boldsymbol {\mu }},,{\boldsymbol {\Sigma }})$, where the mean vector $\bm{\mu}$ and covariance matrix $\bm{\Sigma}=CC^\top$ are randomly generated with entries uniformly sampled from $[-0.5,0.5]$ and ensuring that $\text{cond}(\bm{\Sigma}) \le 50$ for numerical stability. We generate $M=10^5$ i.i.d samples from ${\mathcal {N}}({\boldsymbol {\mu }},\,{\boldsymbol {\Sigma }})$ and aim to compress the dataset while preserving key statistical quantities of interest: the first ($m_1(\bm{x})$) and second ($m_2(\bm{x})$) moments of the data. In this experiment, we found that only 7 correction steps are sufficient to restore the QoIs within the error tolerance of 0.05. Figure~\ref{fig:Multinormal_Phi4} (left) shows significantly reduction in the QoIs error when using our super-resolution method compared to samples obtained by lossy decompression and autoencoders. For a random initialization of the Markov chains, we found that it would take approximately 50 steps to generate samples matched the target QoIs with the same error.

\begin{figure*}
    \centering
    \includegraphics[width=0.99\linewidth]{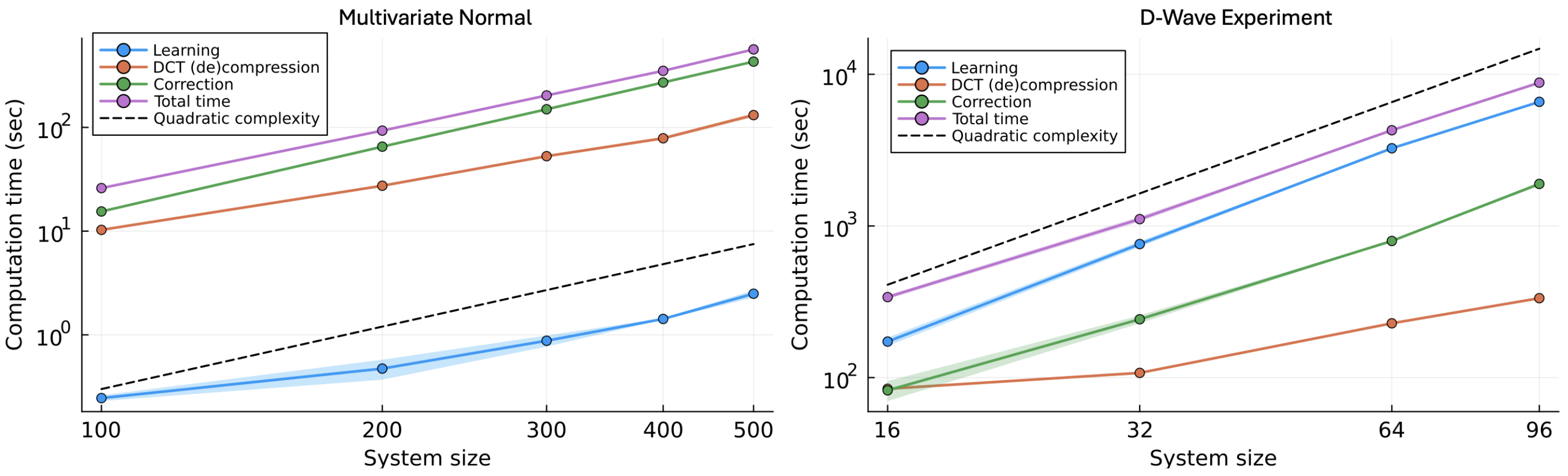}
    \caption{\small Dependence of the wall-clock computation time of our approach on the dimensionality of the data for multivariate Normal distributions (left) and D-wave experiments (right). The figure shows the runtime (in seconds) of each component of our algorithm—learning, DCT-based compression/decompression, and MCMC correction—as a function of system size $N$ for the maximum considered compression level $C=0.9$. We empirically find that all three components scale no worse than quadratically with system size in this case. All data points are averaged over 3 model instances, the error bars are estimated through the empirical standard deviation.}\vspace{-0.2cm}
    \label{fig:Scalability_study}
\end{figure*}

\textbf{Scalar lattice field theory:} We next test our procedure on the scalar $\Phi^4$-theory, a fundamental model in quantum field theory that is widely used to study phase transitions and scalar dynamics in particle and condensed matter physics. We consider the two-dimensional lattice version of the theory on a $16\times16$ grid, re-casted in the dimensionless form $\mathbb{P}(\bm{x}) = \frac{1}{Z} \exp \big( -\alpha \sum_i x_i^4 - \beta \sum_i x_i^2 - \gamma \sum_{(i,j)\in E} x_i x_j \big)$, with parameters $\bm{\theta}=(\alpha,\beta,\gamma)$ sampled uniformly from $[0.1,0.12]\cup[0.2,0.22]\cup[0.3,0.32]$. Using Langevin dynamics with $n=50{,}000$ steps and step size $\Delta T=0.001$, we generate $M=10^5$ i.i.d. samples and verify parameter accuracy through score matching, ensuring $|\bm{\hat{\theta}}-\bm{\theta}|_{\max}\le 0.01$. Our goal is to compress this dataset while preserving three key QoIs: $q_1 = \frac{1}{M} \sum_{s=1}^M \big(\frac{1}{|E|}\sum_{(i,j)\in E} x_i^{(s)} x_j^{(s)} \big)$, $q_2 {=} \frac{1}{M} \sum_{s=1}^M \frac{1}{N}\sum_{i} (x_i^{(s)})^2$, $q_3 {=} \frac{1}{M} \sum_{s=1}^M \frac{1}{N} \sum_{i} (x_i^{(s)})^4$. As shown in Figure~\ref{fig:Multinormal_Phi4} (middle), 20 correction steps with our super-resolution method are sufficient to recover the QoIs within the tolerance of 0.05, whereas random initialization requires at least 200 steps to achieve the same accuracy. Note that the regularized autoencoder performs best for this synthetic benchmark but still maintains a level of error ($\sim 0.15$) above the required error threshold.

\textbf{Molecular dynamics simulation of aluminum:} Finally, we evaluate our method on real scientific data obtained from high-fidelity molecular dynamics (MD) simulations of aluminum using the Embedded-Atom Method (EAM), a widely used many-body potential for modeling metallic interactions. The simulations were carried out with the LAMMPS package \citep{LAMMPS}, a state-of-the-art tool for stress testing and materials design, on a block of aluminum containing $5\times10^5$ particles, constrained in volume and equilibrated at 300 K. Our objective is to compress this dataset while preserving temperature as the key QoI, with
$T = \frac{2\,E_{\mathrm{kin}}}{N_{\mathrm{DOF}}\,k_B}$ where $E_{\mathrm{kin}} = \sum_{i=1}^{N_{\mathrm{atoms}}} \frac{1}{2} m_i v_i^2$ is the total kinetic energy, $N_{\mathrm{DOF}}$ is the number of degrees of freedom, and $k_B$ the Boltzmann constant. In this experiment, 50 correction steps were sufficient to restore the temperature within an error tolerance of 0.5 K. As shown in Figure~\ref{fig:Multinormal_Phi4} (right), our super-resolution method achieves a substantial reduction in QoI error compared to lossy decompression and autoencoder baselines.  

\subsection{Scalability Study}
We examine how the computational complexity of our algorithm scales with the system size for three components of our approach: (1) learning the exponential family distribution from the original data; (2) applying discrete cosine transform (DCT)–based compression and decompression; and (3) correcting the decompressed data using Markov chain Monte Carlo (MCMC) methods guided by the learned model.

\textbf{Synthetic Gaussian data:}
We analyze the case of a multivariate Normal distribution under conditions analogous to the one in Section 3.1, but with the system size $p$ varied from 100 to 500. We choose multivariate Normal distribution for this study due to the easiness of sample generation. For each system size, we generate $M=10^5$ i.i.d. samples from ${\mathcal {N}}({\boldsymbol {\mu }},\,{\boldsymbol {\Sigma }})$. Our objective is to compress this Gaussian dataset to a compression level of 0.9 while preserving key statistical quantities of interest (QoIs)—specifically, the first and second moments—within a tolerance of 0.05. In our experiment, we found that 40 correction steps (using the Langevin dynamics with the step size of 0.05) are sufficient across all tested system sizes to restore the QoIs within the specified tolerance. Notice that learning is the easiest component of the workflow in this specific case because of a simple form of the score matching loss for multivariate Normal distributions, although it is expected to be the most time-consuming element for general distributions.

\textbf{Experimental D-Wave data:} We now test our procedure on real experimental data produced on a D-Wave 2000Q quantum annealer. Each experimental dataset $\mathcal S_i$ consists of $4 \times 10^6$ samples of a 16-qubit system defined by a distinct Ising Hamiltonian. To emulate larger system sizes, we aggregate datasets by independently drawing one sample from each $\mathcal S_{i,i=1,...,k}$ according to its probability distribution and concatenating these into a joint sample of size $16k$. Repeating this procedure $4 \times 10^6$ times yields a dataset $\mathcal S^*$ of $4 \times 10^6$ large samples. As in the Gaussian case, the goal is to compress $\mathcal S^*$ to a compression level of 0.9 while preserving the first $(m_1(\bm{x}))$ and second $(m_2(\bm{x}))$ moments as QoIs within a tolerance of 0.05. In this case, the number of correction steps is observed to scale linearly with system size, approximately at the rate of $10 N$. Figure~\ref{fig:Scalability_study} shows the computation time (in seconds) as a function of system size. Consistent with the Gaussian experiments, all three components of the framework are at most quadratic scaling in this setting. This observed polynomial trend for the super-resolution sampling even at the maximum considered compression rate $C = 0.9$ is encouraging because for non-trivial distributions, the correction sampling from random initializations can take time growing exponentially with the system size \citep{pabbaraju2023provable}.  The number of MCMC correction steps could be reduced at lower compression, offering trade-off between computational efficiency and storage requirements.

\section{Conclusion}
In this work, we addressed a problem for managing the growing data requirements in scientific machine learning. We have introduced a computational framework for scientific data reduction that enables super-resolution sampling preserving user-defined quantities of interest. Our approach focused on establishing enhanced inference-tractability and predictive capacities of exponential family distributions, as well as their sampling properties while guaranteeing the conservation of quantities of interest. Throughout various examples in continuous and discrete domains, we have shown that the proposed approach can achieve significant compression while still accurately recovering key statistical QoIs with minimal correction steps. All codes and data used to produce results in this work are provided as a self-contained archive in the Supplementary Materials. 

\section*{Code Availability}
The code used in this work is publicly available at
\href{https://github.com/lanl-ansi/SuperResolution}{github.com/lanl-ansi/SuperResolution} \cite{2025LFT}.\\

\section*{Acknowledgements}
The authors acknowledge support from the U.S. Department of Energy/Office of Science Advanced Scientific Computing Research Program and from the Laboratory Directed Research and Development program of Los Alamos National Laboratory under Project No. 20230338ER.


\bibliographystyle{plainnat}
\bibliography{references}


\clearpage
\appendix

\thispagestyle{empty}

\onecolumn
\aistatstitle{Scientific Data Compression and Super-Resolution Sampling: Supplementary Materials}

\section{Experimental Details}
We now provide additional details of our the experimental setup and super-resolution sampling workflow. 

\textbf{Ising model:}
 Our procedure to compress the dataset while preserving key statistical QoIs is as follows. First, we compute the first and second empirical moments of the samples, i.e., $m_1(\bm{\sigma})= \frac{1}{M}\sum_{m=1}^M\bm{\sigma}^{(m)}$ and $m_2(\bm{\sigma}) = \frac{1}{M-1}\sum_{m=1}^M (\bm{\sigma}^{(m)} - m_1(\bm{\sigma})) (\bm{\sigma}^{(m)} - m_1(\bm{\sigma}))^\top$, and store them as the target QoIs. Next, we learn a compact representation of the data by learning an Ising model using GRISE. For each node $i \in V$,  we solve a convex optimization problem to estimate the local interaction parameters $\{\hat{J}_{ij},\hat{H}_i\}$. We use the Ipopt solver implemented through the open-source library GraphicalModelLearning.jl to obtain these parameter estimates, although any convex optimization method would be suitable due to the convexity of the objective.

Once the model is learned, we apply the DCT compression procedure to compress the original samples and store them in the compressed form. We use different values of $E_{\text{presv}} \in [0.1,0.3,...,0.9]$, representing varied levels of compression, to analyze the effects of progressive compression on the proposed approach's performance. For reconstruction, we use the proposed procedure, combining naive decompression with learned model-based MCMC correction. Specifically, we decompress our stored data and use each sample to initialize a Markov chain governed by Glauber dynamics and the learned model. These chains (one per sample) are run until the moments of the reconstructed data match the stored QoIs within an error tolerance of 0.05. The error is defined as the maximum element-wise error of the first and second moments of reconstructed samples, i.e., $E = \max \{e_1,e_2\}$, where $e_1 = \|m_1(\bm{\sigma}) - m_1(\bm{\tilde{\sigma}})\|_{\infty}$, $e_2 = \|m_2(\bm{\sigma}) - m_2(\bm{\tilde{\sigma}})\|_{\infty}$, and $\bm{\tilde{\sigma}}$ are the reconstructed samples. Finally, we repeat the experiment and evaluate the performance of our approach using Total Variation (TV) distance, instead of using the QoIs, as the error metric, which is computed as $d_{\mathrm{TV}}(P, Q) = \frac{1}{2} \sum_{\bm{\sigma} \in \Omega} \big| P(\bm{\sigma}) - Q(\bm{\sigma}) \big|$, where $P$ and $Q$ are the original and reconstructed data distribution, respectively.

\textbf{D-Wave analog quantum computer:}
We test our approach on $4\times 10^6$ samples collected from D-Wave 2000Q platform on 16 qubits defined on a chip topology known as Chimera graph. Here, the testing setup and our procedure are analogical to the synthetic case of Ising models, where we aim to compress the data set while aiming to preserve the first and second-order correlations as QoIs.

\textbf{Multivariate Normal Distribution:} To compress the Gaussian dataset and preserve its key statistical QoIs, similar to previous examples with discrete data, we first compute the first ($m_1(\bm{x})$) and second ($m_2(\bm{x})$) empirical moments of the samples, and store them as the target QoIs. Next, we rewrite the density of the normal distribution in our standard form $\mathbb{P}(\bm{x}) = \frac{1}{Z} \exp \Big(-\frac{1}{2}(\bm{x}-\bm{\mu})^\top \bm{\Sigma} (\bm{x}-\bm{\mu})\Big) = \frac{1}{\tilde{Z}} \exp \Big(\bm{x}^\top A \bm{x} + b^\top \bm{x} \Big)$,
where $A = -\frac{1}{2}\bm{\Sigma}^{-1}, b = \bm{\Sigma}^{-1}\bm{\mu}$,
and learn the exponential distribution using the score-matching technique. The score-matching object is solved using gradient descent with 1000 steps and 0.05 step size. Once the model is learned, we apply the DCT compression procedure to compress the original samples. Under the super-resolution recovery, we decompress our stored data and use each sample to initialize a Markov chain governed by Langevin dynamics (with 0.05 step size) and the learned model. These chains (one per sample) are run until the moments of the reconstructed data match the stored QoIs within a tolerance of 0.05.

\textbf{Scalar Lattice Field Theory:}
Our testing setup and procedure are identical to the mutltivariate Normal setting above.

\textbf{Molecular dynamics simulation of aluminum:} 
In this experiment, the system under study consists of a block of aluminum containing $5 \times 10^5$ atoms, simulated under constant volume conditions and equilibrated at 300 K. Each atom is recorded by its position, velocity, and forces, $\bm{x} = [x, y, z, vx, vy, vz, fx, fy, fz]$. The primary quantity of interest (QoI) is the system temperature, defined as  
$T = \frac{2E_{\mathrm{kin}}}{N_{\mathrm{DOF}} k_B}$, 
where 
$E_{\mathrm{kin}} = \sum_{i=1}^{N_{\mathrm{atoms}}} \tfrac{1}{2} m_i v_i^2, N_{\mathrm{DOF}} = n_{\mathrm{dim}} N_{\mathrm{atoms}} - n_{\mathrm{dim}} - N_{\mathrm{fixDOFs}}$, where $E_{\mathrm{kin}}$ is the total kinetic energy of the group of atoms, $m_i=26.982/(6.022\times 10^{23})$,
$n_{\mathrm{dim}}=3$ is the dimensionality of the simulation, 
$N_{\mathrm{atoms}}$ is the number of atoms in the group, 
$N_{\mathrm{fixDOFs}}$ is the number of degrees of freedom removed by fix commands, 
$k_B = 1.38\times 10^{-23}$ is the Boltzmann constant, and $T$ is the resulting computed temperature.

In our approach, we learn the model $\mathbb{P}(\bm{x}) \propto e^{\theta \, \text{QoI}(\bm{x})}$ using score-matching technique (with Ipopt solver and Julia JuMP), store the compressed dataset and then apply our super-resolution recovery strategy using Langevin dynamics (with 0.05 step size) and the learned model. These chains (one per sample) are run until the moments of the reconstructed data match the stored QoIs within a tolerance of 0.5 K. 

\textbf{Autoencoder Comparison:}
Autoencoders are commonly used to learn compact representations of high-dimensional data and to uncover latent structures \citep{hinton2006reducing}. Regularized variants further improve representation learning by incorporating prior knowledge and structural constraints \citep{jarrett2020target,gille2022semi}. We benchmark our framework against both standard and regularized autoencoder baselines.

The autoencoder is $D_{w_1}(E_{w_2}(\bm{x}))$, where $\bm{z} = E_{w_2}(\bm{x})$ denotes the encoder output (or latent variables) and $\hat{\bm{x}} = D_{w_1}(\bm{z})$ the reconstructed sample, with $w_1, w_2$ denoting the respective network parameters. Both encoder and decoder are implemented as standard multilayer perceptrons (MLPs). The autoencoder is trained by minimizing the reconstruction error, $\min_{w_1,w_2} \frac{1}{M} \sum_{s=1}^M \|\bm{x}^{(s)}-D_{w_1}(E_{w_2}(\bm{x}^{(s)}))\|_2^2$.
The regularized autoencoder is trained using the augmented objective,
$ \min_{w_1,w_2} \frac{1}{M} \sum_{s=1}^M \|\bm{x}^{(s)}-D_{w_1}(E_{w_2}(\bm{x}^{(s)}))\|_2^2 + \lambda\text{QoIs}(\bm{x}^{(s)})$,
where $\lambda > 0$ controls the strength of the regularization, and the QoIs are defined consistently with our method. Compression is quantified as the ratio between latent and original dimensions, $C := |\bm{z}|/|\bm{x}|$.

\section{Scalability Study Details}
\label{sec:Appendix_Scalability}

We present additional details for the scalability study of the proposed framework.

\textbf{Synthetic Gaussian Data:}
To ensure numerical consistency across system sizes in this study, we construct a well-conditioned covariance matrix ${\boldsymbol{\Sigma}}$ whose elements do not scale with the dimensionality of the data. We achieve this by first generating a random orthogonal matrix $\mathbf{Q}$ via QR decomposition. Then, we create a diagonal matrix $\mathbf{D}$ with eigenvalues uniformly distributed between 1 and $1/\kappa$, where $\kappa=10$ is the desired condition number. The covariance matrix is then computed as ${\boldsymbol{\Sigma}} = \mathbf{Q} \mathbf{D} \mathbf{Q}^\top$, ensuring that it remains well-conditioned regardless of system size.

Similar to the Gaussian example, we compute the first moment, $m_1(\bm{x})$, and the second moment, $C = m_2(\bm{x})$, of the data, and store them as the target quantities of interest (QoIs). Then, we learn the density of the multivariate normal distribution $\mathbb{P}(\bm{x}) = \frac{1}{Z} \exp \Big(-\frac{1}{2}(\bm{x}-\bm{\mu})^\top \bm{\Sigma}^{-1} (\bm{x}-\bm{\mu})\Big)$,
where $\bm{\mu}$ is estimated using the empirical mean, $\bm{\hat{\mu}} = m_1(\bm{x})$. To estimate $\Theta \approx \bm{\Sigma}^{-1}$, we apply score matching to the centered variable $y = \bm{x} - \bm{\hat{\mu}}$, assuming the form
$p(y) =\frac{1}{Z} \exp\left( -\frac{1}{2} y^\top \Theta y \right).$
This leads to the optimization of the score-matching objective 
$J(\Theta) =  \frac{1}{2}\text{tr}(\Theta^\top \Theta \,C) - \text{tr}(\Theta)$,
which is minimized via gradient descent, with the gradient $\nabla_{\Theta} J = C\Theta - I$, 1000 steps and a step size of 0.05.
Once the model is learned, the remaining DCT-based compression and super-resolution recovery are carried out similarly to the Gaussian example.

\textbf{Experimental D-Wave Data:}
Here, to preserve QoIs of the data, we compute and store the first moment, $m_1(\bm{x})$, and the second moment, $C = m_2(\bm{x})$, of the data. We then learn an Ising model representation of the data using GRISE, solving a convex optimization problem to estimate parameters $\{\hat{J}_{ij}, \hat{H}_i\}$ for each node $i \in V$. The optimization is performed via gradient descent with a step size of 0.25, terminating when improvements fall below $10^{-6}$. Once the model is learned, we compress the original samples using the DCT-based procedure at a compression level of 0.9. 

During super-resolution recovery, we decompress our stored data and use each sample to initialize a Markov chain governed by Glauber dynamics and the learned model. These chains (one per sample) are run until the moments of the reconstructed data match the stored QoIs within a tolerance of 0.05. To restore the QoIs within tolerance 0.05, the number of correction steps is observed to scale linearly with system size, approximately at the rate of $10 N$.

\end{document}